\documentclass{article}

\usepackage{PRIMEarxiv}

\usepackage[utf8]{inputenc} % allow utf-8 input
\usepackage[T1]{fontenc}    % use 8-bit T1 fonts
\usepackage{hyperref}       % hyperlinks
\usepackage{url}            % simple URL typesetting
\usepackage{booktabs}       % professional-quality tables
\usepackage{amsfonts}       % blackboard math symbols
\usepackage{nicefrac}       % compact symbols for 1/2, etc.
\usepackage{microtype}      % microtypography
\usepackage{lipsum}
\usepackage{amsmath}        % math
\usepackage{fancyhdr}       % header
\usepackage{graphicx}       % graphics
\usepackage{nicefrac}       % inline fractions
\usepackage{afterpage}
\usepackage{appendix}
\graphicspath{{media/}}     % organize your images and other figures under media/ folder
\usepackage{multirow}
\usepackage{wrapfig,lipsum,booktabs}
\usepackage{caption}
\usepackage{subcaption}
\usepackage[table,xcdraw]{xcolor}
\usepackage[justification=centering]{caption}

\usepackage[table,xcdraw]{xcolor}
\usepackage[normalem]{ulem}
\useunder{\uline}{\ul}{}
\newcommand{\blue}[1]{\color[HTML]{00009B} {#1}}
\title{Depth Anything in Medical Images:\\ A Comparative Study} 

\author{
  John J. Han, Ayberk Acar, Callahan Henry, Jie Ying Wu \\
  Department of Computer Science \\
  Vanderbilt University \\ 
  Nashville, TN, USA\\
  \texttt{\{john.j.han, ayberk.acar, callahan.g.henry, jieying.wu\}@vanderbilt.edu} \\
}

\begin{document}
\maketitle

\begin{abstract}
Monocular depth estimation (MDE) is a critical component of many medical tracking and mapping algorithms, particularly from endoscopic or laparoscopic video. However, because ground truth depth maps cannot be acquired from real patient data, supervised learning is not a viable approach to predict depth maps for medical scenes. Although self-supervised learning for MDE has recently gained attention, the outputs are difficult to evaluate reliably and each MDE's generalizability to other patients and anatomies is limited. This work evaluates the zero-shot performance of the newly released Depth Anything Model~\cite{depthanything} on medical endoscopic and laparoscopic scenes. We compare the accuracy and inference speeds of Depth Anything with other MDE models trained on general scenes as well as in-domain models trained on endoscopic data. Our findings show that although the zero-shot capability of Depth Anything is quite impressive, it is not necessarily better than other models in both speed and performance. We hope that this study can spark further research in employing foundation models for MDE in medical scenes. 
\end{abstract}


\section{Introduction}
Foundation models are artificial intelligence (AI) networks that learn from significant amounts of a wide variety of different (typically unlabeled) data, making them suitable for many downstream tasks~\cite{li2023multimodal}. Due to the immense size of these models, organizations with access to proper computing resources train these models and release them for the public to use or fine-tune. Technical and research advances in natural language processing (NLP) and computer vision (CV) have enabled multimodal AI applications for various consumer and academic applications~\cite{bommasani2022opportunities}. 

The Depth Anything Model (DAM)~\cite{depthanything} is a recently released foundation model for monocular depth estimation (MDE), exhibiting state-of-the-art zero-shot, few-shot, and in-domain results for both indoor and outdoor scenes. It is a vision transformer-based architecture~\cite{visiontransformers}, whose dataset includes a total of around 63.5 million images (1.5M labeled and 62M unlabeled). The details of training and architecture of Depth Anything are provided in Section~\ref{depthanything_details}. 

Recently, MDE has been a growing research area in medical computer vision, specifically in endoscopic data~\cite{schmidt2023tracking}. By predicting per-pixel depth values, MDEs can enable novel next-generation technologies in medicine, such as intraoperative surgical navigation systems through simultaneous localization and mapping (SLAM)~\cite{chen2019slam} or radiation-free 3D imaging modalities through Structure-from-Motion \cite{liu2020reconstructing, liu2022sage}. However, training MDE models in the medical domain poses unique challenges; ground truth depth maps are impossible to acquire in clinical endoscopic videos, and public medical datasets are scarce. Some datasets~\cite{ozyoruk2020endoslam, c3vd} were released, but they contain videos and depth maps from synthetic phantoms or rely on renderings from anatomical mesh models, both of which are often unrealistic compared to real patient data. As a result, self-supervised learning (SSL) methods have gained popularity in the literature for real endoscopic video, but they often have poor generalization to other patients and cannot be evaluated reliably due to lack of ground truth~\cite{liu2019dense, SHAO2022102338, LIU2023107619}. 

This study compares and evaluates a variety of depth estimators: MiDaS~\cite{birkl2023midas}, ZoeDepth~\cite{zoedepth}, EndoSfM~\cite{ozyoruk2020endoslam}, Endo-Depth~\cite{recasens2021endo}, and Depth Anything~\cite{depthanything} on two well-known medical datasets: EndoSLAM~\cite{OZYORUK2021102058} and rectified Hamlyn Dataset~\cite{recasens2021endo}. We demonstrate that the zero-shot performance of various MDE models trained on general scenes is comparable to that of models directly trained on clinical images. Furthermore, we compare the inference speeds of these models to encourage their use in real-time systems such as SLAM. We hope that this study will motivate further research in this area to actualize novel vision-based surgical interventions.

\section{Related Work and Background}
\subsection{Endoscopic Monocular Depth Estimation}
Monocular depth estimation from an endoscopic video is a relatively new field; it is mostly employed as an intermediary step for 3D reconstruction such as Structure from Motion (SfM) ~\cite{schmidt2023tracking}. Some of the earliest works rely on renderings from synthetic colon models to generate ground truth and use a convolutional neural network with conditional random fields for depth regression~\cite{MAHMOOD2018230}. Chen et al. use a Generative Adversarial Network (GAN) to generate depth maps, which are fused to create 3D reconstructions of the colon~\cite{chen2019slam}. Endo-SfMLearner learns disparity maps between consecutive frames using spatial attention layers~\cite{OZYORUK2021102058}. Although these methods were generally successful on simulated data, relying on phantom models and synthetic renderings for direct supervision is not ideal due to the lack of generalizability to real endoscopic images~\cite{jeong2021depth}. As a result, self-supervised methods are the current dominant paradigm in endoscopic MDE. A popular method is to predict relative camera poses between frames and use warped depth maps for view-consistent outputs. Liu et al. run COLMAP on the endoscopic image sequence (i.e. sparse 3D points and camera poses) to use as a sparse supervisory signal~\cite{Liu_2018}. A Siamese network is used on pairs of frames for dense depth regression~\cite{Liu_2018}. Shao et al. introduce an illumination-invariant optical flow component, called appearance flow, for robust ego-motion and depth estimation~\cite{SHAO2022102338}. Although self-supervised learning methods do not require ground truth depth maps, the resulting networks typically do not generalize well to other anatomies and patients. On the contrary, the scale and reliable transfer learning capabilities of foundation models show promise to overcome these limitations.

\subsection{Foundation Models in Medicine}
Over the last couple of years, extremely capable foundation models were introduced~\cite{zhou2023comprehensive} and sparked various studies in fine-tuning and prompting for medical applications~\cite{lee2024foundation, foun_medicine}. ChatGPT, a large language Model (LLM), was a notable catalyst in using foundation models in medicine due to its ease of use and prompting~\cite{chatgpt}. Some of its applications include automated patient text data interpretation, medical writing, and clinical decision making~\cite{LI2024108013, Lohleader, wefwefwef}. Segment Anything Model (SAM), whose name inspired Depth Anything, is a general-purpose segmentation model that uses a variety of prompting and initialization~\cite{kirillov2023segment}. Its applications in medicine have primarily been developed for cell imaging (e.g. whole slide ~\cite{deng2023segment}) and general medical imaging (\cite{MAZUROWSKI2023102918}). 

As an MDE foundation model, the Depth Anything Model has the potential to succeed in the medical domain. We evaluate zero-shot results with other general-scene models (MiDaS~\cite{birkl2023midas} and ZoeDepth~\cite{zoedepth}) and endoscopic MDE models (Endo-Depth~\cite{recasens2021endo}, EndoSfM~\cite{ozyoruk2020endoslam}, IsoNRSfM~\cite{isonrsfm}) on existing public endoscopic datasets (EndoSLAM~\cite{ozyoruk2020endoslam} and Hamlyn~\cite{recasens2021endo}). We also compare these models' inference speeds on a single-GPU system. 

\subsection{Depth Estimation Models in Medicine}

In this section, we motivate the use of monocular depth estimation in surgical computer-assisted interventions and emphasize the importance of a foundation model in this domain. 

In many endoscopic and laparoscopic surgeries where the surgeon solely relies on a monocular camera, localization and spatial awareness with respect to the anatomy are challenging due to constricted view, simultaneous tool/camera manipulation, and constrained anatomy. To combat this issue, surgeons often use preoperative imaging or navigational systems for guidance and reference. However, in surgeries such as kidney stone removal, the anatomy is no longer spatially consistent with the preoperative imaging due to dilation or shift~\cite{renal}. In addition, intraoperative navigation systems are often expensive and difficult to incorporate~\cite{9374417}. As a result, computer vision algorithms are essential to enable sophisticated guidance systems that can seamlessly integrate into clinical workflow. 

Depth estimation is an essential component in many vision-based navigational and 3D reconstruction methods in the literature~\cite{liu2020reconstructing, liu2019dense, ALI2023129}. Using computer vision to regress pixel-wise depth values can provide intraoperative depth perception without the addition of any sensors or hardware installations. In addition, image-depth pairs in the endoscopic video can strengthen vision-based 3D reconstruction and localization methods, since depth maps give 3D information about each pixel in the frame~\cite{rgbdslam}. This can enable both vision-based postoperative surgical feedback through SfM ~\cite{HONG201422} or real-time intraoperative guidance through SLAM~\cite{liu2022sage}. 

MDE in general scenes is a mature field of literature, with extensive public datasets of both indoor and outdoor scenes. The field is still improving with new state-of-the-art models being developed each year~\cite{birkl2023midas, zoedepth, Ranftl2022, Ranftl2021, ke2023repurposing}. On the contrary, MDE in the clinical domain is relatively new due to the lack of public datasets with real endoscopic data. SSL methods have shown promise to actualize monocular depth estimation from real endoscopic video sequences. 

Although SSL methods are promising avenues of endoscopic MDE, they have some limitations; the lack of ground truth makes it difficult for true evaluation, and it cannot be generalized to other anatomies and patients outside of its training set. To circumvent these issues, a foundation model trained on millions of images may provide a viable solution to medical MDE. This study is dedicated to evaluating and comparing the newly released Depth Anything Model to other MDE models, both in-domain of medical videos and models trained on general scenes. 

\subsection{Depth Anything Model} \label{depthanything_details}
% \subsection{Depth Anything Model} \label{depthanything_details}
The main contribution of Depth Anything is using a relatively small set of labeled images (1.5M) for initial training to create a teacher model, which can generate pseudo labels to train a student network on a larger set of unlabeled images (62M). In the first stage, the authors reproduced the work of MiDaS \cite{birkl2023midas} with pre-trained encoder weights from \cite{dinov2}. Via supervised learning, the model learns a teacher network $T$, which can then generate noisy labels for a student network $S$. The authors found that training the student network in the same configuration as the teacher network's training did not improve performance; as a result, a harder learning objective was imposed via extreme color distortion, Gaussian blurring, and most importantly CutMix~\cite{yun2019cutmix}. For further robustness against noisy labels, $f_i$, the feature vector produced by the encoder of $S$, is aligned with $f_i'$, the feature vector of DINOv2~\cite{dinov2} via the loss term 

$$ L_{feat} = 1 - \frac{1}{HW}\sum_i^{HW}\text{cos}(f_i, f_i')$$

where $(H,W)$ are the image dimensions. This value is discarded if it exceeds threshold $\alpha =0.15$. The authors show that cosine similarity improves their results since a forced projection to $f_i'$ (for example, $L1$ distance) would lead this loss term to dominate the early training process. Furthermore, the thresholding is important to encourage the model to produce finer details in the output depth map. Because DINOv2 is normally used for semantic image tasks like image retrieval and segmentation, the authors noticed decreasing the influence of DINOv2 with a higher threshold $\alpha$ improves the model's performance. Finally, DPT~\cite{Ranftl2021} decodes the feature vector to regress the depth prediction. 

\section{Experiments and Results}
\begin{figure}
\centering
  \includegraphics[width=0.8\linewidth]{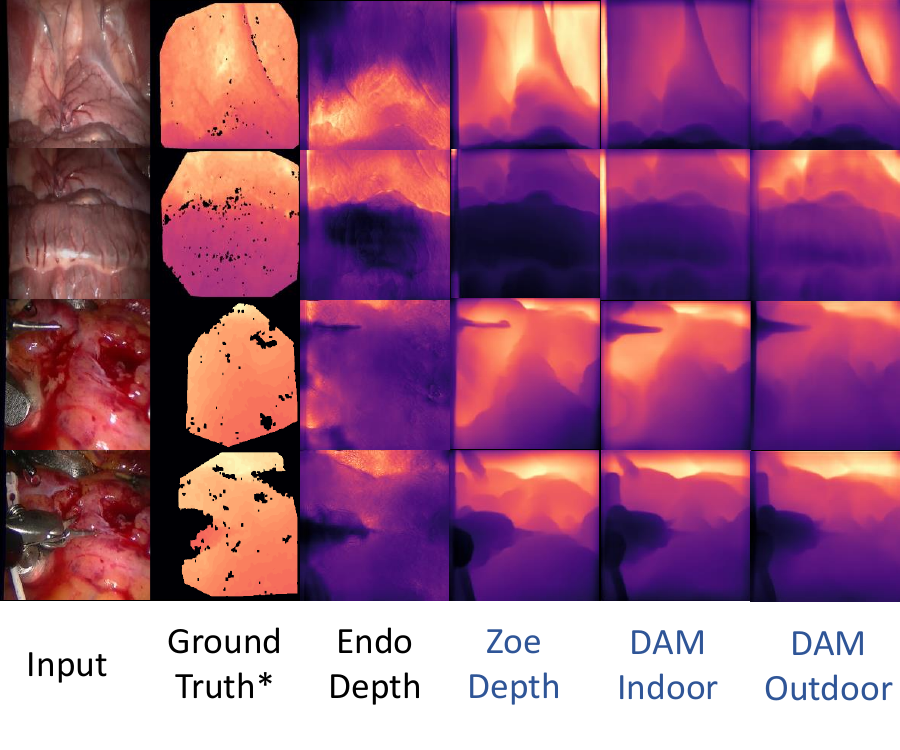}
  \caption{Qualitative results on rectified Hamlyn dataset. Ground truth is created using Libelas software stereo matching~\cite{recasens2021endo}. Model names written in blue are zero-shot models. The monocular+stereo model is used for EndoDepth's visualizations.}
  \label{fig:hamlyn}
\end{figure}

Qualitative results are displayed in Fig.~\ref{fig:endoslam} for EndoSLAM and Fig.~\ref{fig:hamlyn} for Hamlyn Rectified, visually comparing the results of our models with the ground truth. Metrics to quantitatively evaluate the results are given as follows. Given a pixel $d_i$ from the predicted depth map $D$ and its corresponding pixel $\hat{d}_i$from the ground truth depth map $\hat{D}$, we define the Absolute Relative Error (Abs. Rel.), Squared Relative Error (Sq. Rel.), Root Mean Squared Error (RMSE), and $\delta_1$ as the following:

$$ \text{Abs. Rel.} = \frac{1}{N}\sum_i |\hat{d}_i - d_i|/\hat{d_i}$$
$$ \text{Sq. Rel} = \frac{1}{N}\sum_i (\hat{d}_i - d_i)^2/\hat{d_i}$$
$$ \text{RMSE} = \sqrt{\frac{1}{N}\sum_i (\hat{d}_i - d_i^2}$$
$$ \delta_1 = n(\text{max}(\frac{\hat{D}}{D}, \frac{D}{\hat{D}}) < 1.25)$$

where $n(\text{condition})$ is the percentage of pixels that satisfy its argument. 

\afterpage{
\begin{figure}
\centering
  \includegraphics[width=0.9\linewidth]{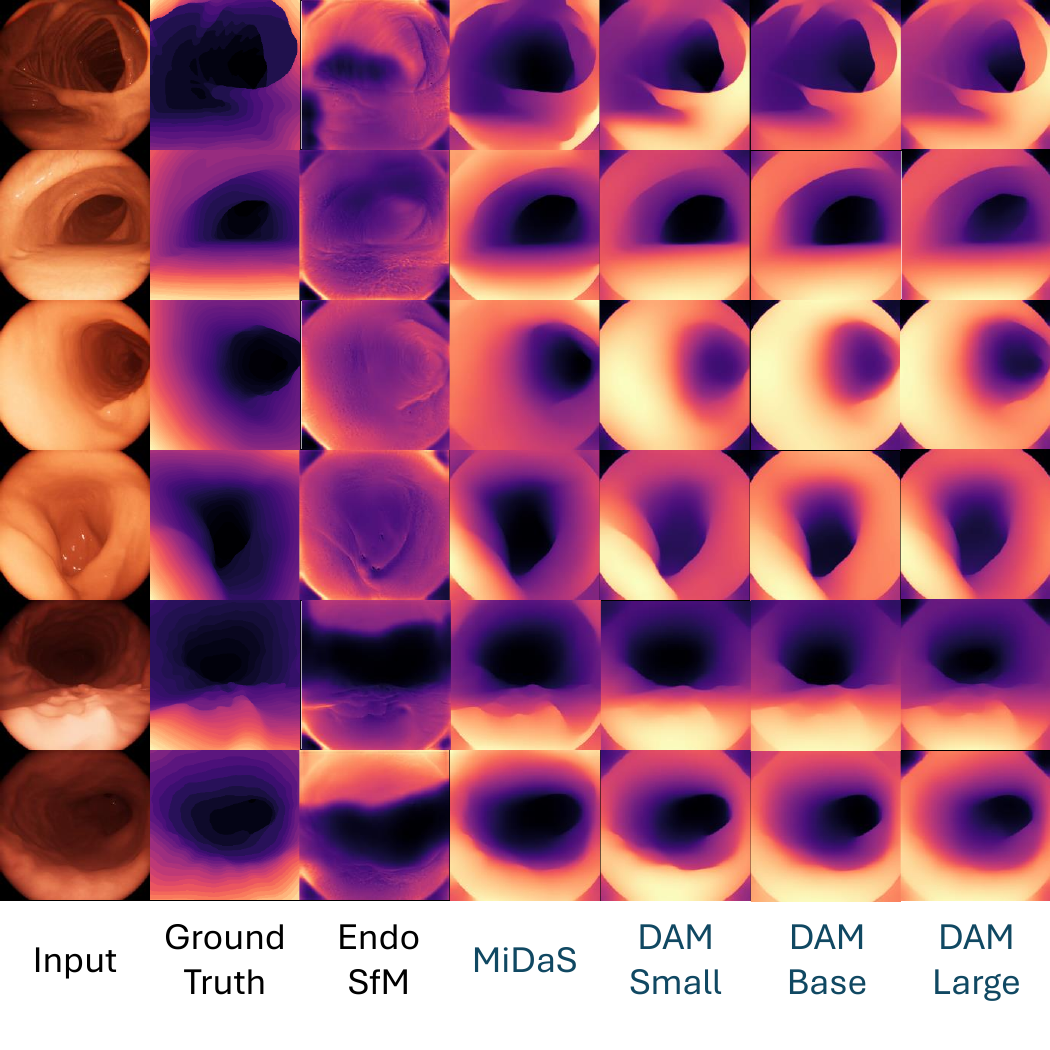}
  \caption{Qualitative results on EndoSLAM. From top to bottom, there are two images each for colon, small intestine, and stomach. Model names written in blue are zero-shot models.}
  \label{fig:endoslam}
\end{figure}

\clearpage
}

\subsection{Evaluation on EndoSLAM Dataset}
The EndoSLAM dataset \cite{ozyoruk2020endoslam} contains a total of 36,023 image-depth pairs across 3 anatomies: stomach, small intestine, and colon. We compare the three Depth Anything Models (small, base, large) with MiDaS~\cite{birkl2023midas}, a depth estimator trained on general scenes, and Endo-SfMLearner~\cite{ozyoruk2020endoslam}, the baseline method of EndoSLAM. We chose these models to compare the performance of zero-shot general MDE models and a model trained on in-domain data. Because monocular depth predictions $D$ do not necessarily have the same scale as their ground truth images $\hat{D}$, we employ the following scaling method to the raw output of each model

$$ D_{scaled} = D \times \frac{\text{median}(\hat{D})}{\text{median}(D)}$$

following previous work. Table~\ref{endoslam_table} displays the results of Depth Anything, MiDaS, and Endo-SfMLearner on each anatomy from EndoSLAM. We implemented Endo-SfMLearner ~\cite{OZYORUK2021102058} using their GitHub repository and evaluated all models with the same method. We note that contrarily to the original EndoSLAM paper that uses minimum/maximum values for scaling, predictions are scaled with the ratio of ground truth and prediction median values to be consistent with other methods.

For the stomach dataset, we observe that the large model of Depth Anything has the best performance across most metrics, despite the fact that it is not directly in-domain. Similarly, MiDaS has the best performance in the colon dataset. Endo-SfMLearner generally has the best performance in the small intestine dataset, achieving the best or second best in all metrics. It is worth noting that DAM-Large does not always outperform DAM-Base in both small intestine and colon sequences. 

\newpage
\begin{center}
% Please add the following required packages to your document preamble:
% \usepackage{multirow}
% \usepackage[table,xcdraw]{xcolor}
% Beamer presentation requires \usepackage{colortbl} instead of \usepackage[table,xcdraw]{xcolor}
% \usepackage[normalem]{ulem}
% \useunder{\uline}{\ul}{}
\begin{table}[]
\centering
% \begin{tabular}{m{2.5cm} m{4cm} m{2.5cm} m{2.5cm} m{2.5cm} m{3cm}}
\begin{tabular}{m{2.5cm}m{4cm}c>{\centering}m{3cm}c}
\toprule
 &
  Method &
  Stomach &
  Small Intestine &
  Colon  \\\midrule 
\multicolumn{1}{c}{} &
  Endo-SfMLearner~\cite{OZYORUK2021102058} &
  0.438 &
  \uline{0.474} &
  \uline{0.551}  \\
\multicolumn{1}{c}{} &
  {\color[HTML]{00009B} MiDaS}~\cite{birkl2023midas} &
  {\color[HTML]{00009B} 0.309} &
  {\color[HTML]{00009B} \textbf{0.351}}&
  {\color[HTML]{00009B} \textbf{0.354}}  \\
\multicolumn{1}{c}{} &
  {\color[HTML]{00009B} DAM-Small}~\cite{depthanything} &
  {\color[HTML]{00009B} 0.354} &
  {\color[HTML]{00009B} 0.585} &
  {\color[HTML]{00009B} 0.616}  \\
\multicolumn{1}{c}{} &
  {\color[HTML]{00009B} DAM-Base}~\cite{depthanything} &
  {\color[HTML]{00009B} \uline{0.284}} &
  {\color[HTML]{00009B} 0.567} &
  {\color[HTML]{00009B} 0.607}  \\
\multirow{-5}{*}{Abs. Rel.$(\downarrow)$} &
  {\color[HTML]{00009B} DAM-Large}~\cite{depthanything} &
  {\color[HTML]{00009B} \textbf{0.239}} &
  {\color[HTML]{00009B} 0.574} &
  {\color[HTML]{00009B} 0.626}  \\\midrule 
 &
  Endo-SfMLearner~\cite{OZYORUK2021102058} &
  3.025 &
  \textbf{8.274} &
  \uline{4.190}  \\
\multicolumn{1}{c}{} &
  {\color[HTML]{00009B} MiDaS}~\cite{birkl2023midas} &
  {\color[HTML]{00009B} 2.961} &
  {\color[HTML]{00009B} 13.921} &
  {\color[HTML]{00009B} \textbf{3.074}}  \\
\multicolumn{1}{c}{} &
  {\color[HTML]{00009B} DAM-Small}~\cite{depthanything} &
  {\color[HTML]{00009B} 2.638} &
  {\color[HTML]{00009B} 12.885} &
  {\color[HTML]{00009B} 5.297} \\
\multicolumn{1}{c}{} &
  {\color[HTML]{00009B} DAM-Base}~\cite{depthanything} &
  {\color[HTML]{00009B} \uline{2.021}} &
  {\color[HTML]{00009B} \uline{12.221}} &
  {\color[HTML]{00009B} 5.273}  \\
\multirow{-5}{*}{Sq. Rel.$(\downarrow)$} &
  {\color[HTML]{00009B} DAM-Large}~\cite{depthanything} &
  {\color[HTML]{00009B} \textbf{1.637}} &
  {\color[HTML]{00009B} 12.653} &
  {\color[HTML]{00009B} 5.629}  \\\midrule 
 &
   Endo-SfMLearner~\cite{OZYORUK2021102058} &
  6.387 &
  \uline{14.787} &
  \uline{6.657}  \\
\multicolumn{1}{c}{} &
  {\color[HTML]{00009B} MiDaS}~\cite{birkl2023midas} &
  {\color[HTML]{00009B} \textbf{5.533}} &
  {\color[HTML]{00009B} \textbf{13.252}}&
  {\color[HTML]{00009B} \textbf{5.502}}  \\
\multicolumn{1}{c}{} &
  {\color[HTML]{00009B} DAM-Small}~\cite{depthanything} &
  {\color[HTML]{00009B} 6.907} &
  {\color[HTML]{00009B} 18.434} &
  {\color[HTML]{00009B} 7.536}  \\
\multicolumn{1}{c}{} &
  {\color[HTML]{00009B} DAM-Base}~\cite{depthanything} &
  {\color[HTML]{00009B} 6.072} &
  {\color[HTML]{00009B} 18.278} &
  {\color[HTML]{00009B} 7.417}  \\
\multirow{-5}{*}{RMSE $(\downarrow)$} &
  {\color[HTML]{00009B} DAM-Large}~\cite{depthanything} &
  {\color[HTML]{00009B} \uline{5.620}} &
  {\color[HTML]{00009B} 18.286} &
  {\color[HTML]{00009B} 7.730}  \\\midrule 
 & 
  Endo-SfMLearner~\cite{OZYORUK2021102058} &
  0.372 &
  \textbf{0.406} &
  \uline{0.354}  \\
\multicolumn{1}{c}{} &
  {\color[HTML]{00009B} MiDaS}~\cite{birkl2023midas} &
  {\color[HTML]{00009B} 0.479} &
  {\color[HTML]{00009B} 0.200} &
  {\color[HTML]{00009B} \textbf{0.484}}  \\
\multicolumn{1}{c}{} &
  {\color[HTML]{00009B} DAM-Small}~\cite{depthanything} &
  {\color[HTML]{00009B} 0.416} &
  {\color[HTML]{00009B} 0.306} &
  {\color[HTML]{00009B} 0.285}  \\
\multicolumn{1}{c}{} &
  {\color[HTML]{00009B} DAM-Base}~\cite{depthanything} &
  {\color[HTML]{00009B} \uline{0.506}} &
  {\color[HTML]{00009B} 0.310} &
  {\color[HTML]{00009B} 0.289}  \\
\multirow{-5}{*}{$\delta_1 (\uparrow)$} &
  {\color[HTML]{00009B} DAM-Large}~\cite{depthanything} &
  {\color[HTML]{00009B} \textbf{0.599}} &
  {\color[HTML]{00009B} \uline{0.317}}&
  {\color[HTML]{00009B} 0.283}  \\\bottomrule \\[0.5ex]
\end{tabular}
\caption{Comparison of four zero-shot and one in-domain methods for relative depth estimation in EndoSLAM dataset. Bold values show the best and underlined values show the second best performing methods. Blue values show the performance for zero-shot estimation.}
\label{endoslam_table}
\end{table}

\end{center}

\subsection{Evaluation on Hamlyn Dataset}
Recansens et. al. rectified the original Hamlyn dataset~\cite{hamlynoriginal} to include paired ground truth~\cite{recasens2021endo}. However, because the depth maps were generated via stereo matching~\cite{libelas}, its ground truth images contain holes and blank areas. Following \cite{recasens2021endo}, we compare the results from sequences 1, 4, 19, and 20. The values of the first five models in Table \ref{hamlyntable} are taken from their manuscript directly. We use the same metrics mentioned above and this time compare Depth Anything's metric depth estimation models (indoor and outdoor) and ZoeDepth \cite{zoedepth}, a high-performing zero-shot metric depth estimator with in-domain baselines. Note that LapDepth~\cite{lapdepth} and IsoNRSfM~\cite{isonrsfm} were trained on \cite{colondepth} to generate these results. For this evaluation, we employ the per-image scaling method $D_{scaled} = D \times \text{median}(\nicefrac{\hat{D}}{D}) $, following~\cite{recasens2021endo}. 

\begin{center}
\begin{table}
    \centering 
    \begin{subfigure}[b]{0.45\linewidth}
    \centering 
        \begin{tabular}{lcccc}\\\toprule 
            Method & Abs. Rel. & Sq. Rel. & RMSE & $\delta_1$ \\ \midrule 
            LapDepth~\cite{lapdepth} & 0.504 & 29.132 & 20.710 & 0.692 \\ 
            IsoNRSfM~\cite{isonrsfm} & 0.097 & 2.563 & 19.815 & 0.930 \\ 
            ED-Mono~\cite{recasens2021endo} & 0.083 & 26.497 & 34.382 & 0.741 \\ 
            ED-Stereo~\cite{recasens2021endo} & 0.083 & 1.379 & \textbf{9.361} & 0.926 \\ 
            ED-M+S~\cite{recasens2021endo} & 0.213 & 11.822 & 23.553 & 0.762 \\ 
            \blue{ZoeDepth}~\cite{zoedepth} & \blue{\underline{0.071}} & \blue{\textbf{1.088}} & \blue{13.100} & \blue{\textbf{0.962}} \\ 
            \blue{DAM-I}~\cite{depthanything} & \blue{\textbf{0.069}} & \blue{\underline{1.097}} & \blue{\underline{12.960}} & \blue{\underline{0.960}} \\ 
            \blue{DAM-O}~\cite{depthanything} & \blue{0.127} & \blue{3.234} & \blue{17.953} & \blue{0.868} \\ \bottomrule
        \end{tabular}
    \caption{Sequence 1}
    \end{subfigure}
    % \hfill
    \hspace{3.5em}
    \begin{subfigure}[b]{0.45\linewidth}
    \centering 
        \begin{tabular}{lcccc}\\\toprule 
            Method & Abs. Rel. & Sq. Rel. & RMSE & $\delta_1$ \\ \midrule 
            LapDepth~\cite{lapdepth} & 0.432 & 12.182 & 11.742 & 0.668 \\ 
            IsoNRSfM~\cite{isonrsfm} & 0.048 & 0.185 & 2.698 & 0.997 \\ 
            ED-Mono~\cite{recasens2021endo} & 0.054 & 0.204 & 2.937 & 0.997 \\ 
            ED-Stereo~\cite{recasens2021endo} & \underline{0.024} & \textbf{0.051} & \underline{1.432} & \textbf{1.000} \\ 
            ED-M+S~\cite{recasens2021endo} & \textbf{0.023} & \underline{0.052} & \textbf{1.428} & \textbf{1.000} \\ 
            \blue{ZoeDepth}~\cite{zoedepth} & \blue{0.079} & \blue{0.453} & \blue{4.427} & \blue{0.977} \\ 
            \blue{DAM-I}~\cite{depthanything} & \blue{0.068} & \blue{0.383} & \blue{4.165} & \blue{0.982} \\ 
            \blue{DAM-O}~\cite{depthanything} & \blue{0.155} & \blue{1.957} & \blue{9.498} & \blue{0.781} \\ \bottomrule
        \end{tabular}
    \caption{Sequence 4}
    \end{subfigure}
    % \hfill 
    \begin{subfigure}[b]{0.45\linewidth}
    \centering 
        \begin{tabular}{lcccc}\\\toprule 
            Method & Abs. Rel. & Sq. Rel. & RMSE & $\delta_1$ \\ \midrule 
            LapDepth~\cite{lapdepth} & 1.234 & 75.260 & 26.742 & 0.493 \\ 
            IsoNRSfM~\cite{isonrsfm} & \textbf{0.062} & \textbf{0.769} & \textbf{6.478} & \textbf{0.960} \\ 
            ED-Mono~\cite{recasens2021endo} & 0.235 & 7.432 & 18.640 & 0.595 \\ 
            ED-Stereo~\cite{recasens2021endo} & \underline{0.169} & 5.275 & \underline{16.213} & \underline{0.796} \\ 
            ED-M+S~\cite{recasens2021endo} & 0.284 & 11.768 & 24.026 & 0.689 \\ 
            \blue{ZoeDepth}~\cite{zoedepth} & \blue{0.192} & \blue{4.670} & \blue{17.793} & \blue{0.706} \\ 
            \blue{DAM-I}~\cite{depthanything} & \blue{0.203} & \blue{5.076} & \blue{18.530} & \blue{0.672} \\ 
            \blue{DAM-O}~\cite{depthanything} & \blue{0.174} & \blue{\underline{4.171}} & \blue{16.758} & \blue{0.741} \\ \bottomrule
        \end{tabular}
    \caption{Sequence 19}
    \end{subfigure}
    % \hfill
    \hspace{3.5em}
    \begin{subfigure}[b]{0.45\linewidth}
    \centering 
        \begin{tabular}{lcccc}\\\toprule 
            Method & Abs. Rel. & Sq. Rel. & RMSE & $\delta_1$ \\ \midrule 
            LapDepth~\cite{lapdepth} & 0.847 & 39.121 & 15.599 & 0.649 \\ 
            IsoNRSfM~\cite{isonrsfm} & \textbf{0.064} & \textbf{0.578} & 6.420 & \textbf{0.988} \\ 
            ED-Mono~\cite{recasens2021endo} & 0.120 & 1.487 & 8.354 & 0.799 \\ 
            ED-Stereo~\cite{recasens2021endo} & 0.120 & 1.212 & 7.491 & 0.845 \\ 
            ED-M+S~\cite{recasens2021endo} & 0.087 & \underline{0.603} & \textbf{5.401} & \underline{0.959} \\ 
            \blue{ZoeDepth}~\cite{zoedepth} & \blue{0.083} & \blue{0.682} & \blue{5.750} & \blue{0.949} \\ 
            \blue{DAM-I}~\cite{depthanything} & \blue{\underline{0.078}} & \blue{0.645} & \blue{\underline{5.564}} & \blue{0.953} \\ 
            \blue{DAM-O}~\cite{depthanything} & \blue{0.092} & \blue{0.846} & \blue{6.436} & \blue{0.909} \\ \bottomrule
        \end{tabular}
    \caption{Sequence 20}
    \end{subfigure}
\caption{Comparison of two zero-shot and five in-domain methods for metric depth estimation in Hamlyn Dataset. Bold values show the best and underlined values show the second best performing methods. Blue values show the performance for zero-shot estimation. Values for LapDepth~\cite{lapdepth}, IsoNRSfM~\cite{isonrsfm}, Endo-Depth (ED)~\cite{recasens2021endo} models are taken from Endo-Depth~\cite{recasens2021endo} paper directly.}
\label{hamlyntable}
\end{table}
\end{center}

We observe a large variance in best performers throughout the four sequences. Generally speaking, ZoeDepth and Depth Anything at the very minimum have comparable results with the baselines and perform best out of all models from sequence 1. Finally, we show the inference speeds of our models on an Intel i9-13900K CPU and NVIDIA GeForce 4090 GPU with image input sizes of (320, 320) in Table~\ref{inferencespeed}. 

\begin{center}
\begin{table}[ht]
\centering
\begin{tabular}{c c c }\\\toprule 
\multicolumn{1}{c}{Models} & \multicolumn{1}{c}{Inference Time (ms)} & \multicolumn{1}{c}{FPS}  \\\midrule
\multicolumn{1}{c}{Endo-SfM~\cite{ozyoruk2020endoslam}} & \multicolumn{1}{c}{7.30} & \multicolumn{1}{c}{136.99}\\
\multicolumn{1}{c}{Endo-Depth~\cite{recasens2021endo}} & \multicolumn{1}{c}{\textbf{4.01}} & \multicolumn{1}{c}{\textbf{249.38}}\\
\multicolumn{1}{c}{ZoeDepth~\cite{zoedepth}} & \multicolumn{1}{c}{82.15} & \multicolumn{1}{c}{12.17}\\
\multicolumn{1}{c}{MiDaS (BEiT-512L)~\cite{birkl2023midas}} & \multicolumn{1}{c}{66.37} & \multicolumn{1}{c}{15.07}\\
\multicolumn{1}{c}{DAM-Small~\cite{depthanything}} & \multicolumn{1}{c}{8.31} & \multicolumn{1}{c}{120.34}\\
\multicolumn{1}{c}{DAM-Base~\cite{depthanything}} & \multicolumn{1}{c}{13.17} & \multicolumn{1}{c}{75.93}\\
\multicolumn{1}{c}{DAM-Large~\cite{depthanything}} & \multicolumn{1}{c}{25.98} & \multicolumn{1}{c}{38.49}\\
\multicolumn{1}{c}{DAM-Indoor/Outdoor~\cite{depthanything}} & \multicolumn{1}{c}{52.93} & \multicolumn{1}{c}{18.89} \\\bottomrule \\[0.5ex]
     
\end{tabular}
\centering
\caption{Average inference Time (ms) for image size (320, 320), computed on Intel i9-13900K, NVIDIA GeForce RTX 4090. All three evaluated models of Endo-Depth (Mono, Stereo, Mono+Stereo) had the same speed.}
\label{inferencespeed}
\end{table} 
\end{center}

\section{Discussion and Conclusion}

Although the authors of Depth Anything demonstrate the superiority of their model compared to ZoeDepth and MiDaS, we observe that it is not necessarily the case in medical endoscopic images. For instance, MiDaS performs better on the colon sequences than Depth Anything in the EndoSLAM dataset, and Endo-SfMLearner and MiDaS outperform all three Depth Anything models in the small intestine sequence. However, DAM-Large generally has the highest accuracy for the stomach video sequence. 

The metric results from Table~\ref{hamlyntable} show that ZoeDepth and Depth Anything-Indoor have similar results across all sequences. However, DAM-Indoor performs notably better than DAM-Outdoor, presumably due to the fact that the interiors of anatomy more closely resemble indoor settings. For sequences 4, 19, and 20, other models trained in-domain outperform both ZoeDepth and all Depth Anything models. On the other hand, we note that these results may be affected by the unideal ground truth maps that the Hamlyn dataset provides, shown in Figure~\ref{fig:hamlyn}. 

In terms of inference time, Endo-Depth is the fastest, twice as fast as Depth Anything's smallest model, while having the highest performance in some sequences of the rectified Hamlyn dataset. However, we note that all evaluated models are capable of some sort of real-time system, as the slowest model was ZoeDepth at 12.17 frames per second. 

Considering that MiDaS, ZoeDepth, and Depth Anything were not fine-tuned at all, there is still room for extended evaluation. The next step of this evaluation study is to expand the amount of models and datasets. For instance, C3VD~\cite{c3vd} and ColonDepth~\cite{colondepth} are also publically available datasets with paired ground truth. Marigold~\cite{marigold} was recently released as a diffusion model-based MDE model, which claims to outperform MiDaS and DPT. However, we noticed that its inference time was relatively slow ($\sim$8\,min per image), limiting its applications in real-time endoscopic navigation. 

In addition, fine-tuning Depth Anything, ZoeDepth, and MiDaS to the medical domain will provide an interesting case study. Fine-tuned on all aforementioned medical datasets, how reliable can these models be for vision-based 3D reconstruction and localization methods? Will its performance translate to real patient images well? Our future work will explore these questions. 

To conclude, in this work, we perform an evaluation and comparison study of the new model, Depth Anything~\cite{depthanything} with other depth estimation models on public endoscopic data. We demonstrate that, despite its high performance and inference speed, it is not necessarily better than its predecessors in both in-domain and other MDEs trained on general scenes. However, its inference speed is favorable for many real-time systems. We hope that this study encourages further exploration in employing foundation models for medical MDE.

%Bibliography
\bibliographystyle{unsrt}  
\bibliography{references}

\end{document}